\documentclass[11pt]{article}
\usepackage{acl2016}
\usepackage{times}
\usepackage{latexsym}
\usepackage{url}
\usepackage{graphicx}
\usepackage{amsmath}
\usepackage{amssymb}
\usepackage{booktabs}
\usepackage{graphicx}
\usepackage{multirow}

\aclfinalcopy 

\title{Longitudinal Analysis of Discussion Topics in an Online Breast Cancer Community using Convolutional Neural Networks}

\author{Shaodian Zhang$^1$, Edouard Grave$^1$, Elizabeth Sklar$^2$ \and \textbf{No\'emie Elhadad}$^1$\\
	    $^1$Columbia University, New York, NY, US\\
	    $^2$King's College London, London, UK\\
	    {\tt \{sz2338,noemie.elhadad\}@columbia.edu}\\
	    {\tt edouard.grave@gmail.com,elizabeth.sklar@kcl.ac.uk}
  }
	  
\date{}

\begin{document}

\maketitle

\begin{abstract}
Identifying topics of discussions in online health communities (OHC) is critical to various applications, but can be difficult because topics of OHC content are usually heterogeneous and domain-dependent.
In this paper, we provide a multi-class schema, an annotated dataset, and supervised classifiers based on convolutional neural network (CNN) and other models for the task of classifying discussion topics. We apply the CNN classifier to the most popular breast cancer online community, and carry out a longitudinal analysis to show topic distributions and topic changes throughout members' participation. Our experimental results suggest that CNN outperforms other classifiers in the task of topic classification, and that certain trajectories can be detected with respect to topic changes.
\end{abstract}

\section{Introduction}
The involvement of the Internet in healthcare gives rise to new perspectives in
eHealth \cite{oh2005ehealth} and changes the way patients consume and
contribute health-related information. Traditionally, patients with
life-threatening conditions receive most of the information about their disease
from their care providers. While providers tend to focus on the clinical impact
of the disease and might ignore the impact of the disease on a patient's
emotional wellbeing and daily life \cite{hartzler2011managing}, support groups,
and more recently online health communities (OHCs), can act as a complementary
source of support for patients \cite{davison2000talks}.
In particular, public online health communities such as Breast Cancer Forum \cite{Wang2012,Elhadad2014,Zhang2014}, the CSN network \cite{Portier2013,Qiu2011}, and Facebook groups \cite{Bender2011} are getting increasingly popular among patients, and have produced unprecedented amount of user-generate content which could be valuable resources for studying OHCs.   

There are many challenges in understanding the very large amount of content authored and read by online health community members, however. Some relate to the quality of information, as well as how the information is consumed and integrated by community members into their daily lives and disease management decisions. 
One fundamental content-related task that is important to downstream content analysis is to identify topics of discussions \cite{Biyani2014}. Previous research suggested that topic, along with emotions, are two basic building blocks of content with respect to OHC content \cite{Portier2013}.
In this study, we focus on investigating prevalences and dynamics of discussion topics in a popular online breast cancer forum. The task is challenging because topics discussed in such OHCs are usually heterogeneous and domain-dependent, and can be different from themes in other biomedical content such as clinical notes, as well as those in other types of general-purpose communities such as Facebook. Previously, topic classification has also been a central issue of text mining in general \cite{Blei2003}, but to our best knowledge no studies has been focused on automated and supervised topic modeling for online health communities.

In this paper, our study objectives are (i) to provide an annotation schema for topic classification; (ii) to contribute an annotated dataset of sentences and posts according to the coding schema; (iii) to experiment with different supervised classification tools, including convolutional neural networks, support vector machines, and labeled latent Dirichlet allocation, to automate the annotation process; and (v) to explore the prevalence and dynamics of different discussion topics in the entire breast cancer community and across member with different disease severities.
Specifically, we ask following research questions:
\begin{enumerate}
\item[1] What is the most effective supervised learning tool in classifying topic of discussions in an online health community?
\item[2] What are the most prevalent topics in discussions in the breast cancer forum?
\item[3] Are there any differences of topic foci among patients of different cancer stages?
\item[4] How does the distribution of topics change through time, as members participate longer in the community?
\end{enumerate}

\subsection{Related Work}
Previously, Sharf observed that in an online breast cancer group, topics regarding basic classifications or definitions of tumors and diagnosis are most prevalent, indicating that Internet support was primarily a complementary source of information in early years \cite{Sharf1997}. A variety of themes such as relationship/family issues became popular in online peer discussions later on \cite{Lewallen2014,Owen2004}, but disease specific topics like treatment, diagnosis, and interpretation of lab test results are still most prevalent \cite{civan2007threading,Meier2007,Cappiello2007}. Specific topics of discussion were identified as well. For example, based on content analysis, Meier and colleagues found that the most common topics in 10 cancer mailing lists were about treatment information and how to communicate with healthcare providers \cite{Meier2007}. Owen and colleagues proposed a topic schema which includes seven categories: outcome of cancer treatment, disease status and processes associated with the cancer, healthcare facilities and personnel, medical test and procedures, cancer treatment, physical symptoms and side effects, and description of cancer in the body \cite{Owen2004}. Based on such schema, prevalence of different topics can be quantified to facilitate content analysis of cancer support groups. More recently, relying on quantitative methods, topic modeling is carried out for public OHCs, but in an unsupervised fashion \cite{Portier2013}.

\section{Methods}

\subsection{Source of data and data processing}
Our work was approved by the Columbia University IRB office. We relied on the
discussion board of the publicly available community from \url{breastcancer.org}. The
entire content of the discussion board was collected in January 2015. The
discussion board is organized in distinct forums, each with threads and posts.
The following pre-processing steps were carried out.

For each post, meta-data about the forum and the thread in which it was
authored was kept, along with author and creation date. The content of each
post was pre-processed by (i) removing all non-textual content (e.g.,
substituting emoticon icons with emoticon-related codes); and (ii) identifying
sentence boundaries using the open-source tool NLTK \cite{loper2002nltk}.

\subsection{Creating the topic schema}
To enable reliable and useful annotation of topics, we established a coding
schema of discussion topics through a literature review of information needs in
online health communities, with an emphasis on breast cancer communities
\cite{meier2007cancer,civan2007threading,blank2010differences,skeels2010catalyzing,wen2011diagnosis,bender2013role,kim2013predictors}. Our objectives were (i) to
devise a coding scheme that is both relevant to describing the information
needs of community members as well as applicable to and robust enough for
automatic topic classification; and (ii) to design a coding scheme that can be
applied to characterizing topics of discussion for either an entire post or its
individual sentences. Furthermore, the annotation schema is such that each unit
of annotation can be labeled according to one or more topics. For instance, a
given post, and even a given sentence can simultaneously convey information
about a treatment and the health system.

The coding scheme was developed using an iterative process to reflect the main
topics of discussion of post content. Preliminary coding of 439 sentences
(corresponding to 37 posts) provided the initial categories and guidelines for
coding. Upon review and discussion, infrequently used categories were collapsed
into larger concepts, and the 439 sentences were coded again to verify
sufficient agreement between the two initial coders. The 439 sentences and
their codes were used as training instances for the later coders, along with
the coding guidelines.

Our final topical scheme contains 11 topics, as listed in Table \ref{schema}.
It is noteworthy that the topics focus on informational support, rather than
emotional dimensions and range from clinical to daily matters. 

\begin{table*}
\begin{tabular}{|l|lp{10cm}|}
\hline
Topic&Abbreviation&Description\\
\hline
Alternative&ALTR&alternative and integrative medicine\\
Daily&DAIL&daily cancer-related experience\\
Diagnosis&DIAG&diagnoses, measurements, and results of tests\\
Finding&FIND&health finding, sign, symptom or side effect \\
Health Systems&HSYS&health systems patients interact with, including nurses, doctors, practices, hospitals, and insurance companies\\
Miscellaneous&MISC&greetings, uninformative sentence, or any sentence, which does not fit under any other annotation label\\
Nutrition&NUTR&nutrition\\
Personal&PERS&personal information\\
Resources&RSRC&link, pointer, or quote towards an external information resource\\
Test&TEST&testing procedures (but not results of tests)\\
Treatment&TREA&treatments, including procedures, medications and therapeutic devices\\
\hline
\end{tabular}
\caption{Annotation schema for breast cancer forum text}
\label{schema}
\end{table*}

We also learned from the preliminary coding that members may shift topic of discussion in a post, which reminded us that to achieve better granularity sentence-level coding would be necessary. As such, our manual annotation described below were carried out at sentence level rather than post-level.

\subsection{Manual Annotation}
We selected a subset of posts (1008 posts consisting of 9016 sentences) from the original dataset
described above. The posts were selected from the different forums, where
each forum focuses on specific aspects of breast cancer management, such as
diagnosis and treatment options, support through chemotherapy, nutrition,
alternative treatments, and daily life. Posts were thus grouped in batches of
50 posts per manual annotation session.

Sentences were coded according to double annotation followed by an adjudication
step from one dedicated adjudicator throughout the annotated dataset. Three
coders were hired for the annotation, all female native English speakers with
undergraduate degrees. To train for the annotations, coders practiced
annotating the 439 sentences (37 posts) referred to above using the annotation
guidelines. Inter-annotator agreement with gold-standard topic annotation was
monitored throughout training, and training was terminated when a coder had
achieved a 0.6 Kappa (agreement statistic) with the gold-standard annotation
\cite{cohen1960coefficient}. Note that given the large number of potential
labels in the schema and the fact that each sentence can be labeled according
to multiple topics, this is a particularly stringent training constraint.
Afterwards, each batch of posts was assigned two coders and was doubly
annotated at the sentence level. Finally, the adjudicator went through all
posts, resolved differences between coders and made final decisions over
sentence topic labels.

\subsection{Topic classification}
Because a given sentence in a post can be described according to multiple
topics (e.g., a sentence can be about a treatment, nutrition, and daily matters
all at once), the task of automating the topic coding can be cast as a
multi-label classification: for each sentence, there can be up to N labels,
where N is the number of topics in the schema. This type of classification is
more challenging than single-label classification, where one sentence can be
described by only one label chosen from the N topics in a schema.
Traditionally, there are two approaches for multi-label, multi-class
classification: problem transformation methods and algorithm adaptation methods
\cite{tsoumakas2007multi}. 

In this paper, we rely on three different supervised classifiers, a labeled LDA classfier \cite{Ramage2003}, an SVM \cite{suykens1999least}, and a convolutional neural network \cite{kim2014convolutional}. 
They represent three types of mainstream supervised learning frameworks: generative graphical models, discriminative max-margin linear classifiers, and neural networks. Within these three models, labeled LDA and neural networks are able to handle multi-label classification naturally since they allow multiple outputs. 
For the SVM, we consider N binary, single-label classifiers and aggregates the N outputs into one multi-label. 

For the labeled LDA classifier, we rely on an self-implemented Gibbs sampler for labeled LDA, based on the open source LDA implementation \cite{Heinrich2005}\footnote{http://jgibblda.sourceforge.net}. The two hyper-parameters of the model, alpha and beta, are set as 0.1 and 0.5 experimentally. For SVM, we rely on the open source tool LibSVM \cite{CC01a}, using linear kernel and all default parameters.

The convolutional neural network we used has one hidden convolutional layer.
First, the sequence of words is represented as a sequence of vector of dimension $D = 100$, by using a lookup table.
The word embeddings used in this lookup table were pre-trained, by using the word2vec algorithm, on the entire unannotated dataset from the same forum.
Then we take the convolutions of this sequence of ``word vectors'' with $H$ filters, obtaining a score for each filter and each position in the sentence.
In order to obtain a fixed-size representation of the sentence, we perform max-pooling (over the positions in the sentence).
We finally apply a fully connected layer to obtain a score for each topic.
Since the dataset is imbalanced, we propose to use asymetric costs for positive and negative examples.
The ratio between these costs is denoted by the scalar $\alpha$.
In our experiments, $H$ is set to 800 and $\alpha$ is set to 0.25.

Prior to training the classifiers, the following pre-processing and feature
selection steps were carried out: (1) all the words in the corpus were stemmed;
(2) stopwords were removed from the vocabulary; (3) dimensionality reduction were carried out by doing Named Entity Recognition (using Stanford NER
\cite{finkel2005incorporating}) to recognize Person, Location,
Organization names as well as special tokens such as
number, money, time. 
In addition, to make the comparison across tools more meaningful, we also use the word embedding input of CNN as features for SVM, examining how it differs from bag of words representations.

\subsection{Application to the entire community to support longitudinal analysis}
We applied the best performed classifier on all sentences in the entire unannotated dataset. For each post, we assigned it topic labels that are associated with more than 1/10 of sentences in the post.
As such, based on the aggregated post-level topic labels, we are able to identify 1) what are the most prevalent topics in general in the community; 2) if there are any differences of topics among members of different cancer stages. We did not examine other factors than cancer stage in this study, because cancer stage is one particular profile information that can be accessed.

Armed with topic labels for each post in the dataset, we also conducted following longitudinal analyses to take timestamp into account. The primary objective for our analysis was to assess if participation in the community has an impact on topic of discussion. We thus compared distributions of topics of posts published in different periods of time with respect to user’s registration date, and tracked their changes. As such, each data point is the average frequency of a topic within all posts in a given time slice (e.g., all posts published by their authors after 3 weeks of their joining the community). To show both short-term and long-term changes, three measures of time progression are used (represented as x-axis): post, day, and week. 

\section{Results}

\subsection{Manual annotation}
Table \ref{dataset} shows distributions and example sentences for different topics in the manually-annotated dataset. Treatment and Miscellaneous sentences are the most frequent topics in our annotated dataset, whereas Alternative Medicine and Test topics are the least prevalent.  The high number of Miscellaneous sentences is explained by the fact that most posts start with greetings and end with encouragements, blessings, and signatures (all categorized as Miscellaneous in our coding).

\begin{table*}[htp]
\begin{center}
{
\begin{tabular}{|l|lp{10cm}|}
\hline
\textbf{Topic}  & \textbf{\#Sentences} & \textbf{Example}\\
\hline
ALTR & 302 & I tried everything to no avail \& in desperation had acupuncture.\\
DAIL & 600 & I use virgin organic coconut oil on my skin and all organic cosmetics, shampoo, conditioner, laundry detergent, household cleaner, the works!\\
DIAG & 1127& My cancer was a 1.2 cm mucinous bc in a duct, with low growth rate.\\
FIND & 1195& I don't feel faint or anything- it just feels weird- anyone else out there had this happen?\\
HSYS & 864 & I don't know where you are located, but I would start with the Cancer Treatment Centers of America.\\
MISC & 1956& Hope this helps, cheers\\
NUTR & 608 & I am staying on a bland diet, eating every 2 hours, and forcing fluids, but am worried about tomorrow based on what happened last time.\\
PERS & 1011 & He has a family history of very high triglycerides. \\
RSRC & 568& I just did internet research and here is a good site with information on Curcumin\\
TEST & 295 & When I went in for my second mammogram on Dec. 18th, the radiologist told me I had to go get a biopsy based upon the mammogram.\\
TREA & 2078& I'm just curious about other warriors experience with herceptin.\\
\hline
ALTR,NUTR & 113 & I read that cinnamon capsules could help with lowering glucose and ldl in our blood. \\
HSYS,TREA & 104 & After dealing with the insurance company for weeks.....she finally started taking the Xeloda last month.\\
\hline
\end{tabular}
}
\caption{Topic labels and the number of manually annotated sentences according
  to each topic. For each topic, an example of manually annotated sentence is
  provided. The table also includes two examples with multiple labels.
  } 
\label{dataset}
\end{center}
\end{table*}

Table \ref{kappa} shows the inter-annotator agreement for each pair of
annotators across the three annotators. Among the three coders, the first coder
annotated all 1008 posts, while the other two complimentary coders are assigned
part of the whole data set. The reminder of the paper reports results on the
adjudicated annotation. 

\begin{table}
\begin{center}
\begin{tabular}{|l|c|c|}
\hline
Label & Coder 1 and 2 & Coder 1 and 3\\
\hline
Avg K & 0.50 & 0.62\\
ALTR & 0.36 & 0.29\\
DAIL & 0.30 & 0.50\\
DIAG & 0.50 & 0.71\\
FIND & 0.56 & 0.61\\
HSYS & 0.56 & 0.68\\
MISC & 0.38 & 0.76\\
NUTR & 0.70 & 0.69\\
PERS & 0.13 & 0.61\\
RSRC & 0.63 & 0.58\\
TEST & 0.69 & 0.70\\
TREA & 0.67 & 0.71\\
\hline
\end{tabular}
\caption{Inter-rater agreements between the three topic coders measured by Cohen's Kappa. Note that coder 1 annotated all posts while coder 2 and coder 3 annotated two complimentary parts of the data. Therefore, no agreement is calculated between coder 2 and coder 3. }
\label{kappa}
\end{center}
\end{table}

\subsection{Topic classification}
The classifiers were evaluated in a 5-fold cross validation framework using
precision, recall, and F measure. In order to evaluate the overall performance
of the system across all topics, micro average precision, recall and F are also
calculated \cite{yang1999evaluation}. Micro average takes distribution of
labels into consideration, and it makes more sense in this study because of the
imbalance of labels in the dataset. Experiments with a baseline system are
also carried out, which simply tags every sentence with all possible
labels. Aggregated results for the sentence-level classification
are given in Table \ref{performance}. 

\begin{table}[htp]
\begin{center}
\begin{tabular}{|l|l|l|l|l|l|}
\hline
&bsline & l-lda & svm & svm-e & cnn\\
\hline
Micro & 19.3 & 54.4 &  55.8 & 58.3 & 65.4 \\
\hline
ALTR & 6.5 & 9.2 & 9.4 & 30.7 & 35.5 \\
DAIL & 12.5 & 30.1 & 28.8 & 46.4 & 48.1\\
DIAG & 22.2 & 58.8 & 60.2 & 65.3 & 67.1 \\
FIND & 23.4 & 50.1 & 50.9 & 60.0 & 60.3 \\
HSYS & 17.5 & 45.4 & 41.1 & 55.3& 57.7 \\
MISC &35.7 & 76.2 & 75.8 & 71.4 & 78.1 \\
NUTR & 12.6& 57.3 & 58.6 & 68.4 & 72.8\\
PERS &20.2 & 24.4 & 26.5 & 47.7 & 47.8 \\
RSRC &11.9& 48.0 & 48.3 & 55.2 & 61.1 \\
TEST &6.3 & 27.6 & 26.1 & 47.9 & 52.6\\
TREA &37.5 & 65.7 & 66.0 & 64.2 & 73.6\\
\hline
\end{tabular}
\caption{Topic classification performance measured by F score on different topic categories, with five classifiers. bsline: the system simply tags all sentences with all 11 labels; l-lda: the labeled LDA classifier; svm: the SVM classifier using bag of words as features; svm-e: the SVM classifier using word embedding as features; cnn: the convolutional neural network classifier}
\label{performance}
\end{center}
\end{table}

\subsection{General prevalence of topics}
Prevalence of all topics in the entire forum at post-level is given in Table \ref{all-topics}. 
The most prevalent topic is personal (PERS), with 24.6\% of posts labeled as
such, followed by treatment (TREA, 24.6\%) and diagnosis (DIAG, 9.3\%). The least
prevalent topics are alternative medicine (ALTR, 0.2\%) and test (TEST, 1.0\%).
It is noteworthy that MISC did not show up in post-level annotation, because it is a default category assigned only when no other topics are identified in all sentences of the post. As such, its prevalence is extremely low at post-level and it is not of interest to our following analysis. 

Clinically relevant topics such as treatment, diagnosis, and finding are more prevalent than non-clinical ones across the breast cancer forum, with one exception of PERS. Topic distribution in the entire BC dataset is more skewed that that in the annotated dataset, because the annotated dataset was sampled toward collecting more posts of rare topics such as alternative medicine (ALTR). 

\begin{table}[htp]
\begin{center}
{
\begin{tabular}{|l|l|l|l|l|}
\hline
\textbf{ALTR} & \textbf{DAIL} & \textbf{DIAG} & \textbf{FIND}& \textbf{HSYS}\\
0.2&7.4&9.3&6.3&7.8\\
\hline
\textbf{NUTR} & \textbf{PERS} & \textbf{RSRC} & \textbf{TEST}& \textbf{TREA}\\
3.9&24.9&1.7&1.0&24.6\\
\hline
\end{tabular}
}
\caption{Percentages of all topics at post level, based on automated topic classification. } 
\label{all-topics}
\end{center}
\end{table}

\subsection{Topic prevalence stratified by cancer stage}

In the BC dataset, many users self-reported disease information in profiles, including cancer diagnoses and treatment histories. These profile information show up in signatures when authors post, which is available to the public. In particular, out of all 57,424 authors in the dataset we crawled, 17,950 (31.3\%) have their cancer stage information available in signatures. Among them, 2,325 are stage 0 (total number of posts: 170,610), 5,968 are stage I (total number of posts: 600,500), 5,907 are stage II (total number of posts: 661,990), 2,447 are stage III (total number of posts: 229,955), and 2,438 are stage IV (total number of posts: 460,313).

\begin{figure}
\centering
\includegraphics[scale = 0.3]{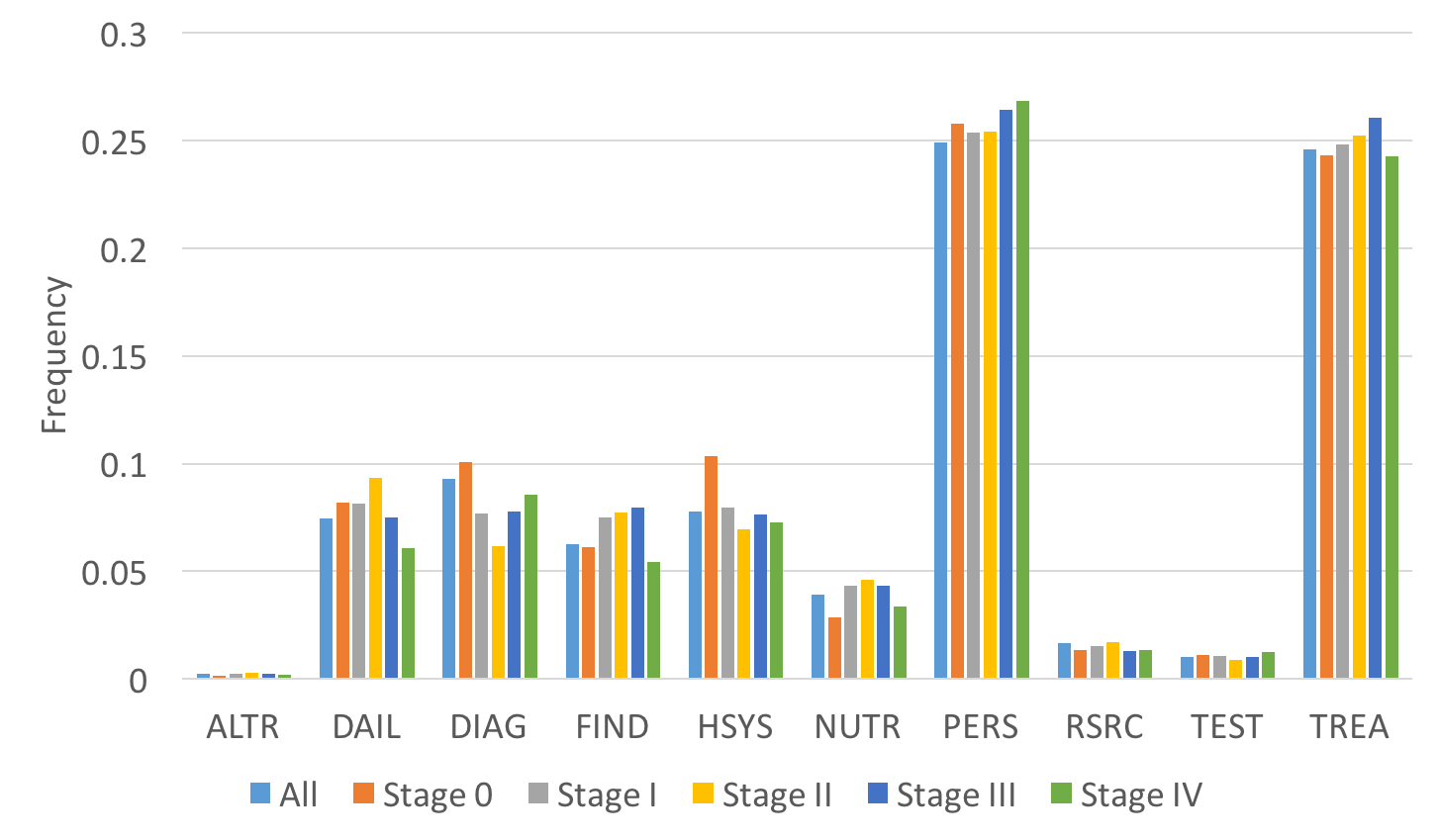}
\caption{Frequencies of topics of posts, stratified by cancer stages of authors. }
\label{fig:topic_distribution_stage}
\end{figure}

Topic distributions of posts published by members of different cancer stages are given in table \ref{fig:topic_distribution_stage}. Statistical tests (multi-variate and univariate) were also carried out between numbers of different stages. Most visible differences in figure \ref{fig:topic_distribution_stage} are statistical significant, given relatively large sample size. 
Stage 0 users focus more on cancer diagnosis and health systems, which are typical topics at early times of cancer journeys. Stage IV members, counter-intuitively, discuss more about personal lives but significantly less about treatment and clinical findings. This seems to suggest that stage IV members rely on the forum to exchange emotional more than informational support with their peers. Another explanation might be that these members are so sick that few treatment options are effective for them.

\subsection{Topic trajectory of users}

\begin{figure*}
\includegraphics[width=16cm]{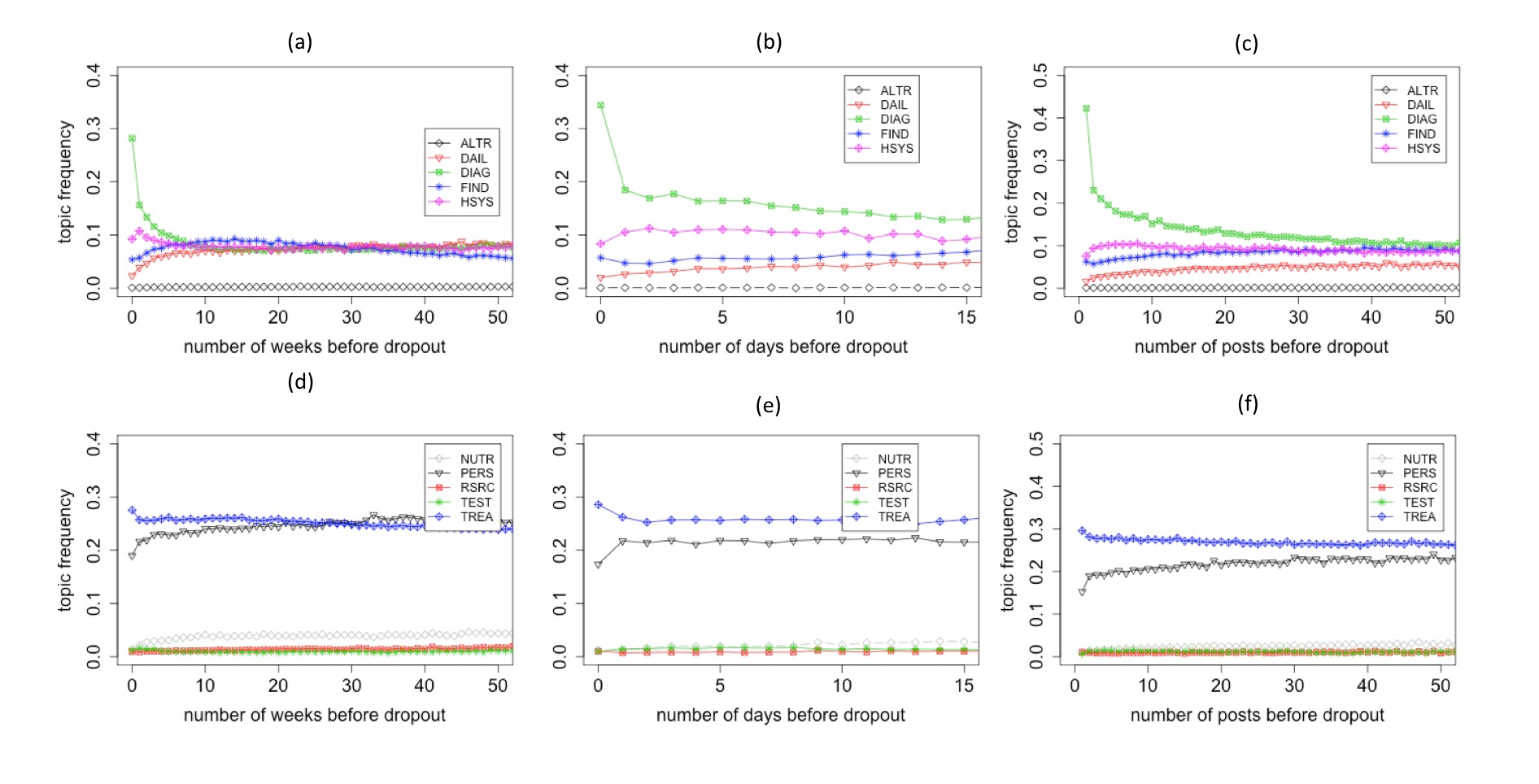}
\caption{How topic frequencies change through time after members join the community. X axes represents the time point after members' first activity. Y axis is the average topic frequency of all posts that are published in the corresponding time. Units of x axes in (a)(d), (b)(e), and (c)(f) are weeks, days, and post orders, respectively.}
\centering
\label{fig:topic_changes}
\end{figure*}

Figure \ref{fig:topic_changes} shows changes of frequencies of topics after members' joining the community, in weeks, days, and individual posts, respectively.
Several types of trajectories are identified. First, diagnosis is the most dominant topic at early stages of participation, especially in first posts and first days. Second, prevalence of some topics such as personal (PERS), daily matters (DAIL), and nutrition (NUTR) grow steadily, while prevalences of diagnosis (DIAG) and treatment (TREA) decline as members stay longer in the community. Third, frequencies of health systems (HSYS) and findings (FIND) increase at the beginning, but slide after reaching the peaks. Finally, alternative medicine (ALTR), laboratory test (TEST), and resources (RSRC) are unpopular topics throughout members' participation. The results seem to suggest that members' focus shifted from informational support, represented by clinically concentrated topics such as diagnosis and treatment, to emotional support, represented by personal focused on topics such as nutrition and daily lives. 

\section{Discussions}
A wide range of topics are discussed in the breast cancer community, ranging from clinically relevant ones such as diagnosis and treatment to more daily matters such as nutritional supplements and personal lives. 
In the breast cancer forum, personal matters and treatment are the most dominant topics, possibly representing a mix of emotional support and informational support being exchanged. 

Cancer stage plays a role in deciding members' topics of discussions. Early stage members, many of whom are newcomers to the community, care more about diagnosis related information. Stage 0 members, in particular, focus on whether certain signs indicate cancer. They also exchange anecdotes about their experiences of visiting healthcare providers when being diagnosed. Late stage members, such as stage IV members, usually stay in the community for longer time as their cancer develop. For these members, seeking information is no longer the primary motivation of participation; on the contrary, they establish closer relationships with their online peers, and disclose more personal information and support each other emotionally. 
It it noteworthy, however, that cancer stage information extracted from signatures may be inaccurate, since members may not report stage change timely. Also, it is naturally the case that members with late stage cancer are more likely to be long time users, which makes length of membership an important confounder in considering differences between members of different stages. 

Finally, we found that members shifted their focus in participation, from clinically relevant topics to more casual topics. This coincides with the difference between cancer stages, and confirms that the difference is at least partly caused by length of participation. As members stay longer in the community and build up closer relationship with their peers, they tend to disclose more personal information, discuss more private stories, and exchange more support emotionally. 
\section{Conclusion}
In this paper, we provide a multi-class schema, an annotated dataset, and supervised classifiers based on convolutional neural network (CNN) and other models for the task of topic classification for online health community text. We apply the classifier on the most popular breast cancer online community, the discussion boards of breastcancer.org, and carry out longitudinal analysis at scale to show topic distributions and topic changes throughout members' participation. Our experimental results suggest that CNN outperforms other classifiers in the task of topic classification. We also found that although personal and disease related topics are most prevalent, members of different cancer stages have different foci of topics. Finally, members change their interest as they participate, becoming increasingly interested in more personal topics in online discussions.

\section*{Acknowledgement}
This work is supported by National Institute of General Medical Sciences Grant R01GM114355.

\bibliography{acl2016}{}
\bibliographystyle{acl2016}

\end{document}